%% file: aaai.tex
\documentclass[letterpaper]{article} % DO NOT CHANGE THIS
\usepackage{aaai25}  % DO NOT CHANGE THIS
\usepackage{times}  % DO NOT CHANGE THIS
\usepackage{helvet}  % DO NOT CHANGE THIS
\usepackage{courier}  % DO NOT CHANGE THIS
\usepackage[hyphens]{url}  % DO NOT CHANGE THIS
\usepackage{graphicx} % DO NOT CHANGE THIS
\usepackage{natbib}  % DO NOT CHANGE THIS AND DO NOT ADD ANY OPTIONS TO IT
\usepackage{caption} % DO NOT CHANGE THIS AND DO NOT ADD ANY OPTIONS TO IT
\usepackage{algorithm}
\usepackage{algorithmic}

\usepackage{newfloat}
\usepackage{listings}
\DeclareCaptionStyle{ruled}{labelfont=normalfont,labelsep=colon,strut=off} % DO NOT CHANGE THIS
\floatstyle{ruled}
\newfloat{listing}{tb}{lst}{}
\floatname{listing}{Listing}
\usepackage{hyperref}% -- This package is specifically forbidden

\title{SemStereo: Semantic-Constrained Stereo Matching Network for Remote Sensing}
\author{
    Chen Chen\textsuperscript{\rm 1,\rm 2}, Liangjin Zhao\textsuperscript{\rm 1}, Yuanchun He\textsuperscript{\rm 1}, Yingxuan Long\textsuperscript{\rm 1,\rm 2}, \\
    Kaiqiang Chen\textsuperscript{\rm 1}\thanks{Corresponding author.}, Zhirui Wang\textsuperscript{\rm 1}, Yanfeng Hu\textsuperscript{\rm 1}, Xian Sun\textsuperscript{\rm 1}
}
\affiliations{
    \textsuperscript{\rm 1}Key Laboratory of Target Cognition and Application Technology,\\
Aerospace Information Research Institute, Chinese Academy of Sciences\\
    \textsuperscript{\rm 2}University of Chinese Academy of Sciences\\
    \{chenchen235, longyingxuan23\}@mails.ucas.ac.cn, \{zhaolj004896, heyc, chenkq, wangzr, huyf,  sunxian\}@aircas.ac.cn 
}

\usepackage{bibentry}
\usepackage{amssymb}
\usepackage{xcolor}
\usepackage{amsmath}
\usepackage{subcaption}
\usepackage{cleveref}
\usepackage{booktabs} 
\usepackage{multirow}

\begin{document}

\maketitle

\input{sec/0_abstract}
\begin{links}
\link{Code}{https://github.com/chenchen235/SemStereo}
\end{links}
\input{sec/01_intro}
\input{sec/02_related}
\input{sec/03_method}
\input{sec/04_experiment}
\input{sec/05_conclusion}
\section{Acknowledgments}
This work was supported by the National Nature Science Foundation of China under Grant 62331027, and supported by the Strategic Priority Research Program of the Chinese Academy of Sciences, Grant No. XDA0360303.

% \bibliography{aaai}

\end{document}

%% file: sec/0_abstract.tex
\begin{abstract}
% Stereo matching is a challenging task in the context of large-scale urban scenes, where the presence of texture-less regions and boundary poses significant challenges to achieving accurate stereo matching. In this paper, we propose S3GNet, a novel semantic segmentation serialization-guided multi-task framework designed to fully leverage the assistance of semantic segmentation for stereo matching. Firstly, we propose a Shared Siamese U-shape Feature Extractor for robust feature extraction followed by semantic supervision. Secondly, our model incorporates an Inter-task Transfer Module for efficient linking of the two tasks. Last, a Satellite Fast Attention Concatenation Volume is constructed to generate initial disparity maps, followed by our Semantic Selective Refinement Branch for disparity refinement. We conduct experiments on large urban scenes (US3D) and achieve state-of-the-art performance. We also extend our S3GNet to common scenes, demonstrating state-of-the-art accuracy on Scene Flow and competitive performance on KITTI 2015. The source code will be available on github after a double blind review. \textcolor{red}{} and the heights distribution of true object

% Remote sensing semantic stereo 3D reconstruction necessitates the extraction of semantic segmentation contours and precise estimation of height through stereo matching. 
Semantic segmentation and 3D reconstruction are two fundamental tasks in remote sensing, typically treated as separate or loosely coupled tasks.
% Current methods typically employ separate models or multi-task models with limited sharing of shallow features, ignoring the uniqueness of observation perspectives and failing to fully exploit the potential of inter-task consistency.
Despite attempts to integrate them into a unified network, the constraints between the two heterogeneous tasks are not explicitly modeled, since the pioneering studies %usually use only shallow shared features,
% \textcolor{red}{
either utilize a loosely coupled parallel structure or engage in only implicit interactions, 
failing to capture the inherent connections.
%We observe that objects of the same category tend to exhibit similar disparity values due to the limitation of the observation perspective close to the bird's-eye view. 
In this work, we explore the connections between the two tasks and propose a new network that imposes semantic constraints on the stereo matching task, both implicitly and explicitly.
Implicitly, we transform the traditional parallel structure to a new cascade structure termed Semantic-Guided Cascade structure, where the deep features enriched with semantic information are utilized for the computation of initial disparity maps, enhancing semantic guidance.
Explicitly, we propose a Semantic Selective Refinement (SSR) module and a Left-Right Semantic Consistency (LRSC) module.
The SSR refines the initial disparity map under the guidance of the semantic map.
The LRSC ensures semantic consistency between two views via reducing the semantic divergence after transforming the semantic map from one view to the other using the disparity map.
% Consequently, we introduce a Semantic-Guided Cascade structure for implicit consistency by sharing more features, a Semantic Selective Refinement branch for explicit intra-class disparity consistency and a semi-/self-supervised method for Left-Right Semantic Consistency. % to promote task consistency in our approach. 
%Furthermore, due to the high cost of remote sensing semantic labels, existing datasets either do not have, or only have semantic labels for the source view, causing the model to be biased on the view. To address this problem, we propose a semi-/self-supervised method to enforce left-right semantic consistency by obtaining pseudo labels supervised by reference view semantics.
%Furthermore, existing methods typically train models using data from a single view, resulting in model bias in view. To address this concern, we propose a semi-supervised method that enforces Left-Right Semantic Consistency by obtaining pseudo-labels for the right semantic supervision. 
% Experimental results on the US3D (satellite scene) and WHU (aerial scene) dataset demonstrate that our CS$^{2}$Net achieves state-of-the-art performance for both semantic segmentation and stereo matching. 
Experiments on the US3D and WHU datasets demonstrate that our method achieves state-of-the-art performance for both semantic segmentation and stereo matching.
\end{abstract}

%% file: sec/01_intro.tex
\section{Introduction}
\label{sec:intro}

%\input{figs/1}

%\quad In the field of remote sensing for large-scale scenes, rapid and accurate semantic urban 3D reconstruction provides crucial spatial information support for planning, infrastructure development, and emergency response\cite{kadhim2016advances, US3D, kunwar2020large}. 
Semantic segmentation and stereo matching are two underlying tasks towards semantic urban 3D reconstruction, which requires both semantic and 3d details derived from high-resolution remote sensing images~\cite{kadhim2016advances, US3D}.
% Semantic segmentation and stereo matching are usually considered as two separate tasks in semantic 3D city reconstruction, using semantic segmentation models~\cite{long2015fully,chen2017rethinking,zhao2017pyramid,xie2021segformer} to extract target contours and stereo matching models~\cite{gc,gwc,acv,igev,xu2023accurate} to extract height information, respectively.
They are usually considered as two independent tasks due to the inherent domain gap that characterizes each task, using a semantic segmentation network~\cite{xie2021segformer, jing2021psrn, kang2021picoco, kang2022disoptnet} for classification and a stereo matching network for height extraction~\cite{zhang2019ga,zhang2020adaptive,acv,igev,xu2023accurate}, respectively.
Afterwards, post-processing is conducted to fuse the parallel results~\cite{qin2019pairwise, kunwar2020large,sun2024gable}.

%Semantic segmentation and stereo matching are two fundamental tasks towards semantic urban 3D reconstruction, which requires both semantic and 3d details derived from high-resolution remote sensing images~\cite{kadhim2016advances, US3D, kunwar2020large}. 
%These two tasks are usually considered as two unrelated tasks in remote sensing image interpretation \cite{blaschke2010object,li2019deep}, urban reconstruction~\cite{he2021dsm, he2022hmsm}, scene understanding~\cite{long2015fully,chen2017rethinking,zhao2017pyramid,xie2021segformer}.% and autonomous driving~\cite{gc,gwc,acv,igev,xu2023accurate}.

%miss the importance of consistency between the two heterogeneous tasks. 

\begin{figure}[!t]
\centering
\includegraphics[width=\linewidth]{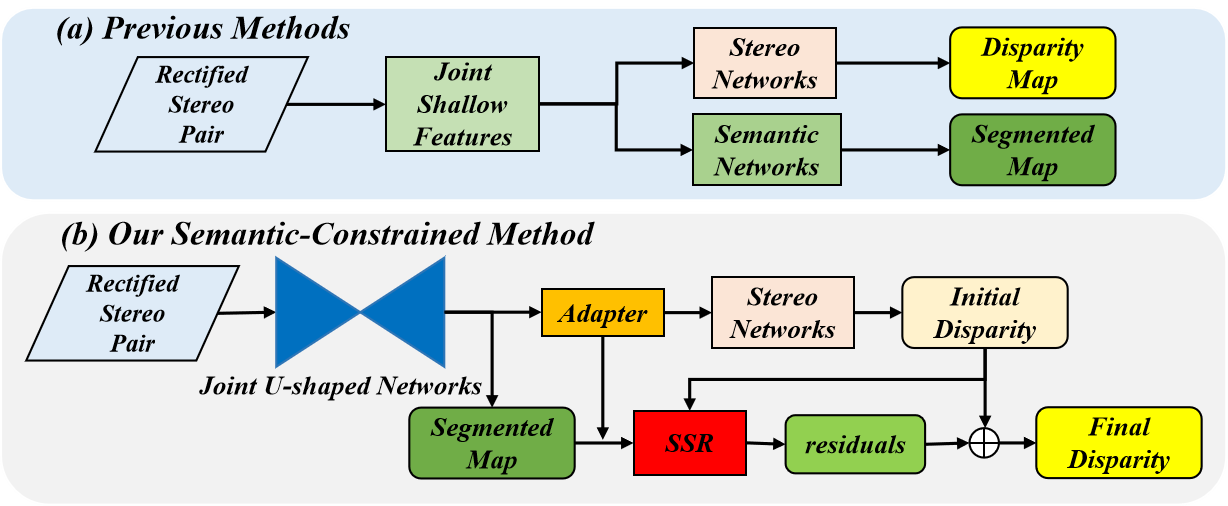}
\caption{\label{fig:framework_comparison}
A comparison between our and previous methods.}
\end{figure}

% Recent attempts~\cite{yang2018segstereo, zhang2019dispsegnet, qin2019pairwise, sspcv, dovesi2020real, kunwar2020large, liao2023s, yang2024s3net} are conducted to integrate these two heterogeneous tasks in a unified framework, resulting in higher accuracy. 
Further studies~\cite{cheng2017segflow, kendall2018multi, song2020edgestereo, liao2023s} attempt to fuse the two heterogeneous tasks in a multi-task network to achieve higher accuracy.
% It is demonstrated that semantic segmentation improves stereo matching in areas with less texture and occlusions~\cite{??}, 
% % while 
% and} stereo matching aids in distinguishing confusing categories in semantic segmentation~\cite{??}. 
% These methods 
The typical multi-task learning methods usually adopt a parallel structure with two branches for semantic segmentation and stereo matching respectively, as illustrated in ~\Cref{fig:framework_comparison}~(a).
Shallow features are shared to establish implicit and weak connections between the distinct tasks~\cite{zhang2019dispsegnet, dovesi2020real, liao2023s}.
Even though it is demonstrated that semantic segmentation improves stereo matching in areas with less texture and occlusions~\cite{yang2018segstereo, sspcv, dovesi2020real}, 
% % while 
and stereo matching aids in distinguishing confusing categories in semantic segmentation~\cite{liao2023s, yang2024s3net}, the underlying mechanisms remain unexplored, leading to a failure to capture the inherent connections between the heterogeneous tasks.

\begin{figure}[!t]
\centering
\includegraphics[width=\linewidth]{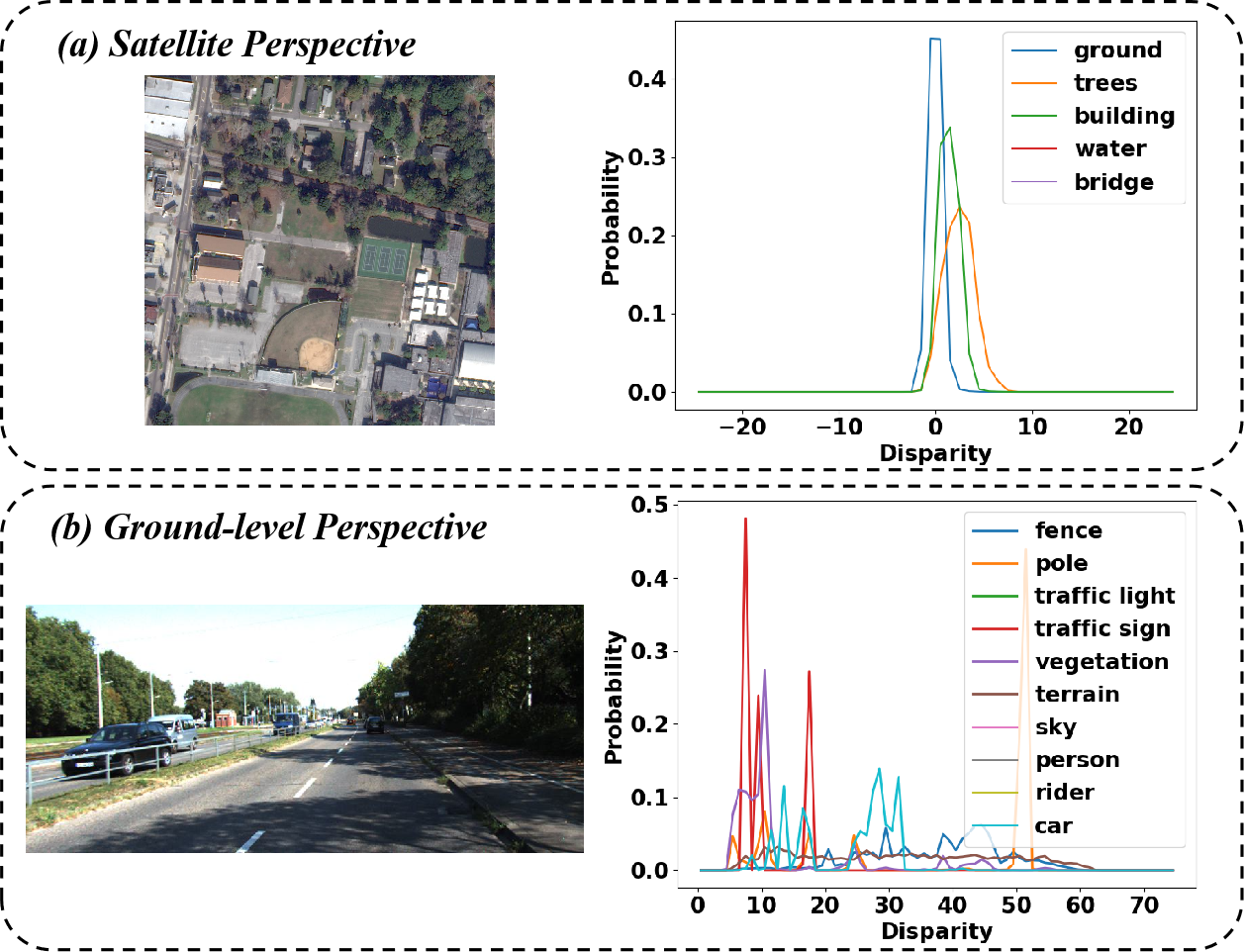}
\caption{
The distribution of disparity per semantic category in satellite and ground-level perspectives, respectively.
% The distribution of disparity per semantic category in (a) satellite \cite{US3D} and (b) ground-level perspectives \cite{menze2015object}, respectively.
% The uniqueness of the observation perspective in remote sensing scenes leads to a single-peak distribution of disparity within the same semantic category, which demonstrates a stronger inter-task consistency compared to the multi-peak distribution in street view scenes.
%Objects within the remote sensing domain~\cite{US3D, IADF_GRSM_201903} of the same category typically exhibit similar disparity values.
%We observe that in remote sensing scenes (US3D), the disparity, which represents the height of real objects, roughly follows a normal distribution for objects of the same category. 
%~\cite{US3D, IADF_GRSM_201903}
%where disparity represents the distance between objects and the camera, the disparity of objects of the same category randomly changes with distance. This difference is attributed to the distinct imaging modalities employed in these domains. 
}
\label{fig:disparity_category}
\end{figure}

%We notice that the disparity of objects with the same category usually has close values as presented in. 
In this work, we aim to uncover the connections between semantic categories and disparities in remote sensing and bridge the task domain gap, with the goal of guiding efficient and interpretable network design for improved accuracy.
As illustrated in ~\Cref{fig:disparity_category}, disparities corresponding to the same category are concentrated within a distinct and narrow range in remote sensing images, while this characteristic does not apply to typical images taken from a ground-level perspective. 
We assume that this might (partially) explain why the heterogeneous tasks can mutually benefit each other.
Based on this assumption, we propose a novel network termed \textit{SemStereo} which imposes the semantic constraints on the stereo matching task via implicitly and explicitly modeling their connections.

% We observe that remote sensing provides a bird's-eye view with negligible perspective effect (size diminishing with distance), resulting in a single-peak disparity distribution for objects of the same category. This contrasts with street view, where the perspective effect leads to a multi-peak form, implying a stronger inter-task consistency in remote sensing (\Cref{fig:intro1}). 
% In this work, we make full use of this consistency both implicitly and explicitly. 

%Inspired by the SSCV-Net of street view scenes \cite{sspcv}, which improves the accuracy of stereo matching by adding a semantic cost volume, but still retains the original spatial cost volume. 
Specifically, we firstly transform the traditional parallel structure (\Cref{fig:framework_comparison}~(a)) into a novel cascade structure termed  Semantic-Guided Cascade (SGC) structure (\Cref{fig:framework_comparison}~(b)).
Instead of only sharing shallow features, we implicitly strengthen the semantic constraints on stereo matching via feeding the deep features right before the segmentation map enriched with semantic information to the stereo network for the initial disparity generation. 
Explicitly, we propose a Semantic Selective Refinement (SSR) module, which uses semantic segmentation maps to guide the disparity maps refinement via learning the residuals for compensation, as illustrated in ~\Cref{fig:framework_comparison}~(b). 
Furthermore, we propose a Left-Right Semantic Consistency (LRSC) module, which explicitly imposes the semantic constraint on stereo matching via reducing the divergence of segmentation maps (or semantic feature maps in the absence of semantic supervision) between the two views after converting one view to another based on the disparity map, as illustrated in ~\Cref{fig:2}~(c).
In summary, we achieve tighter semantic constraints on stereo matching via the implicit deeper semantic feature sharing, the explicit semantic-guided disparity refinement, and the explicit disparity-based cross-view semantic consistency.

% In implicit terms, we transform the traditional parallel structure (Figure~\ref{fig:disparity_category}~(a)) into a new cascade structure termed Semantic-Guided Cascade (SGC) structure (\Cref{fig:intro2} (b)), which obtains the initial disparity map based on the features with richer semantic information in the segmentation decoder feature layers, which deepens the implicit constraints of semantic guidance. 

% In explicit terms, to explicitly leverage the consistency constraints between semantic categories and disparity, we propose the Semantic Selective Refinement (SSR) module (\Cref{fig:intro2} (c)) which uses the class-wise semantic information to correct the initial disparity map via learning the residuals.% from the segmented map.

% Moreover, there is a more general explicit constraint overlooked by previous methods. Since disparity is the pixel difference in the x-axis direction between corresponding points in the left and right views, this should also be maintained between the left and right semantic predictions. To this end, we propose a semi-/self-supervised approach, Left-Right Semantic Consistency (LRSC) supervision, that utilizes the disparities to derive pseudo-labels for the reference (right) view via warping the source (left) semantic labels. 

Our method achieves state-of-the-art performance on the US3D~\cite{IADF_GRSM_201903, le20192019, US3D} and WHU~\cite{liu2020novel} datasets for both semantic segmentation and stereo matching tasks, demonstrating the effectiveness of SemStereo.
Further ablation studies highlight the significance of modeling the connections between semantic categories and disparities.

%% file: sec/02_related.tex
\section{Related Work}

\label{sec:related}

\noindent\textbf{Stereo Matching.}
% Convolutional neural networks have demonstrated significant promise in stereo matching. 
% DispNet \cite{mayer2016large}, a pioneering study, creates the first end-to-end network for stereo matching. 
% \textcolor{blue}{
Cost volume construction is crucial in stereo matching, encompassing methods like correlation volume~\cite{mayer2016large,luo2016efficient}, concatenation volume~\cite{zbontar2015computing,gc} and their combinations~\cite{gwc,shen2021cfnet,acv}.
The correlation volume~\cite{mayer2016large,luo2016efficient} is derived via computing the inner product between the feature volumes of the two views, producing a single-channel map per disparity level depicting the similarity.
Despite its computational efficiency, the collapsed sing-channel map loses the abundant contextual information in the feature volume, reducing accuracy.
% As a pioneering study, DispNet~\cite{mayer2016large} introduced the first end-to-end network for stereo matching, demonstrating the significant potential of convolutional neural networks in this area.
% }
% \textcolor{blue}{
% It employs a vanilla correlation volume that aggregates two thick stacks of feature maps from both views into a single-channel correlation map, resulting in the loss of substantial information needed for accurate disparity estimation, and consequently, reduced accuracy.
% }
% But they employ a correlation volume with single-channel, resulting in loss of information.
% \textcolor{blue}{
The concatenation volume~\cite{gc,zbontar2015computing} addresses this by concatenating the feature volumes from both views~\cite{zbontar2015computing,gc}, but it lacks explicit modeling of similarity measurements.
Further studies~\cite{gwc,shen2021cfnet} aims to combine the strengths of both methods, integrating contextual information and explicitly modeling similarity by concatenating both the correlation volume and concatenation volume.
% }
% \textcolor{blue}{
Nevertheless, these methods~\cite{gc,zbontar2015computing,gwc,shen2021cfnet} suffer from a higher computational cost due to the high dimensions of the cost volumes and the subsequent 3D convolutions.
% }
% Subsequent works have attempted to develop novel cost volumes to enhance the precision, such as a concatenated volume~\cite{gc} to provide abundant content information, a group-wise correlation volume \cite{gwc} with multiple channels to enhance similarity measurements, and a spatial pyramid pooling integrated~\cite{psm} to leverage contextual information. However, these methods~\cite{gc, psm, gwc} require a larger set of parameters for regularization in the subsequent aggregation network. 
% In response, 
DSMNet~\cite{he2021dsm} simplifies 
% \textcolor{blue}{
the regularization network via replacing the regular 3D convolutions with 
factorized 3D convolutions.  
% \textcolor{blue}{
ACVNet~\cite{acv,xu2023accurate} introduces an attention concatenation volume, which learns attention weights based on the correlation volume to filter the concatenation volume, thereby reducing the need for computationally expensive 3D convolutions.
% }
The fast version Fast-ACV~\cite{acv} further enhances efficiency by using the top-k disparity priors from the coarse branch, achieving a better balance between accuracy and speed.
% \textcolor{blue}{
In this work, we use the Fast-ACV to derive the initial disparity map, followed by refinement branches, but our focus is on exploring efficient methods and the potential of improving stereo matching via incorporating semantic constraints in remote sensing. 

\input{figs/2}

\noindent\textbf{Stereo Matching Using Semantic Clues.} 
 %In recent years, significant progress has been made in leveraging semantic clues to augment stereo matching. Segstereo~\cite{yang2018segstereo} incorporates a segmentation sub-network to guide stereo matching with semantic clues, effectively establishing the synergy between semantic segmentation and stereo matching. Dovesi et al. \cite{dovesi2020real} shift their focus towards developing a real-time multi-task model within a lightweight framework, but performed poorly in terms of accuracy. Wu et al.~\cite{sspcv} integrate semantic cost volumes and spatial cost volumes, achieving significant results. For large-urban scenes, some methods~\cite{kunwar2020large,qin2019pairwise} still adopt a parallel framework and share shallow features to process tasks individually. These methods do not exploit the consistency between tasks and the characteristics of large-scale urban scenes, resulting in unsatisfactory results. 
 %Previous works~\cite{gc,psm,gwc,acv,igev,long2015fully,chen2017rethinking,zhao2017pyramid,xie2021segformer} achieve good results in these two separate tasks. 
 %Early works~\cite{bai2016exploiting, behl2017bounding, cheng2017segflow} have shown the benefits of integrating semantic information to improve optical flow estimation and scene flow estimation, demonstrating the powerful auxiliary ability of semantic information in other tasks. 
 % \textcolor{blue}{
 A pioneering study~\cite{hane2013joint} integrates the semantic segmentation results with 3D information through post-processing, based on a category-specific smoothness assumption.
 It is then expanded to larger, higher-resolution oblique aerial scenes~\citep{blaha2016large} using a hierarchical scheme.
 These works are non-deep learning methods, resulting in inferior results.
 Further studies~\cite{qin2019pairwise, kunwar2020large} improve the performance by separately training two separate networks for distinct tasks.
 However, these approaches remain limited to post-processing, failing to fully leverage semantic information for stereo matching.
 % }
 % There have been some previous works on using semantic clues to assist in 3D reconstruction. \cite{hane2013joint} is the pioneer in proposing the category-specific smoothness assumption, using a post-processing method to jointly combine semantic and depth maps, thereby enhancing the quality of reconstruction. %However, their work is confined to non-deep learning approaches and is limited to qualitative analysis on a small scale.
 % \cite{blaha2016large} expands the scope to larger, higher-resolution oblique aerial scenes by proposing a hierarchical scheme. 
 %Their approach involves associating and post-processing point clouds generated by traditional methods with semantic segmentation results to improve reconstruction results. 
 % However, their works are confined to post-processing methods and non-deep learning approaches. Subsequent works \cite{qin2019pairwise, kunwar2020large} adopt a parallel framework where tasks are treated separately by deep networks, achieving better performance. But they still adopt post-process fusion methods, failing to mine implicit relationships between tasks. 
 %Early works show the importance of incorporating semantic information into the network for joint training, such as optical flow estimation ~\cite{bai2016exploiting, cheng2017segflow} and scene flow estimation~\cite{behl2017bounding}. In the realm of stereo matching, to fully exploit the complementary relationship between semantic segmentation and stereo matching in joint training, several fusion methods are proposed. 

% \textcolor{blue}{
An alternative approach involves integrating pre-acquired semantic labels into stereo matching networks by expanding the input dimensions~\cite{US3D}.
However, obtaining these semantic labels is highly time-consuming, which limits practical applications.
Multi-task learning methods~\cite{zhang2019dispsegnet,dovesi2020real} address this limitation by jointly predicting a semantic segmentation map and a stereo map within an end-to-end trainable network.
Experimental analyses reveal that stereo matching benefits from semantic segmentation in textureless~\cite{yang2018segstereo, sspcv, dovesi2020real} and occluded regions, while stereo matching helps to clarify confusing categories for semantic segmentation~\cite{liao2023s}.
% }
% \textcolor{blue}{
These methods typically employ two parallel branches for the two tasks, sharing a shallow feature extractor, which leads to a weak and implicit coupling between them.
A further study~\cite{yang2024s3net} generates one segmentation map and one disparity map from the same cost volume, strengthening the influence of semantic clues on stereo matching.
% }
% \textcolor{blue}{
However, the influence of the semantic clues on the stereo matching remains implicit, and the underlying mechanism is not well understood.
% }
% \textcolor{blue}{
In this work, we dive into the connections between semantic categories and stereo matching from the remote sensing perspective and propose a new framework.
It both implicitly strengthens the impact of semantic segmentation on stereo matching by sharing deep features (i.e., the Semantic-Guided Cascade structure), and explicitly modeling the intra-class disparity consistency (i.e., the Semantic Selective Refinement branch) and semantic consistency between views (i.e., the Left-Right Semantic Consistency supervision).

%% file: figs/2.tex
\begin{figure*}[!htbp]
\centering
\includegraphics[width=\linewidth]{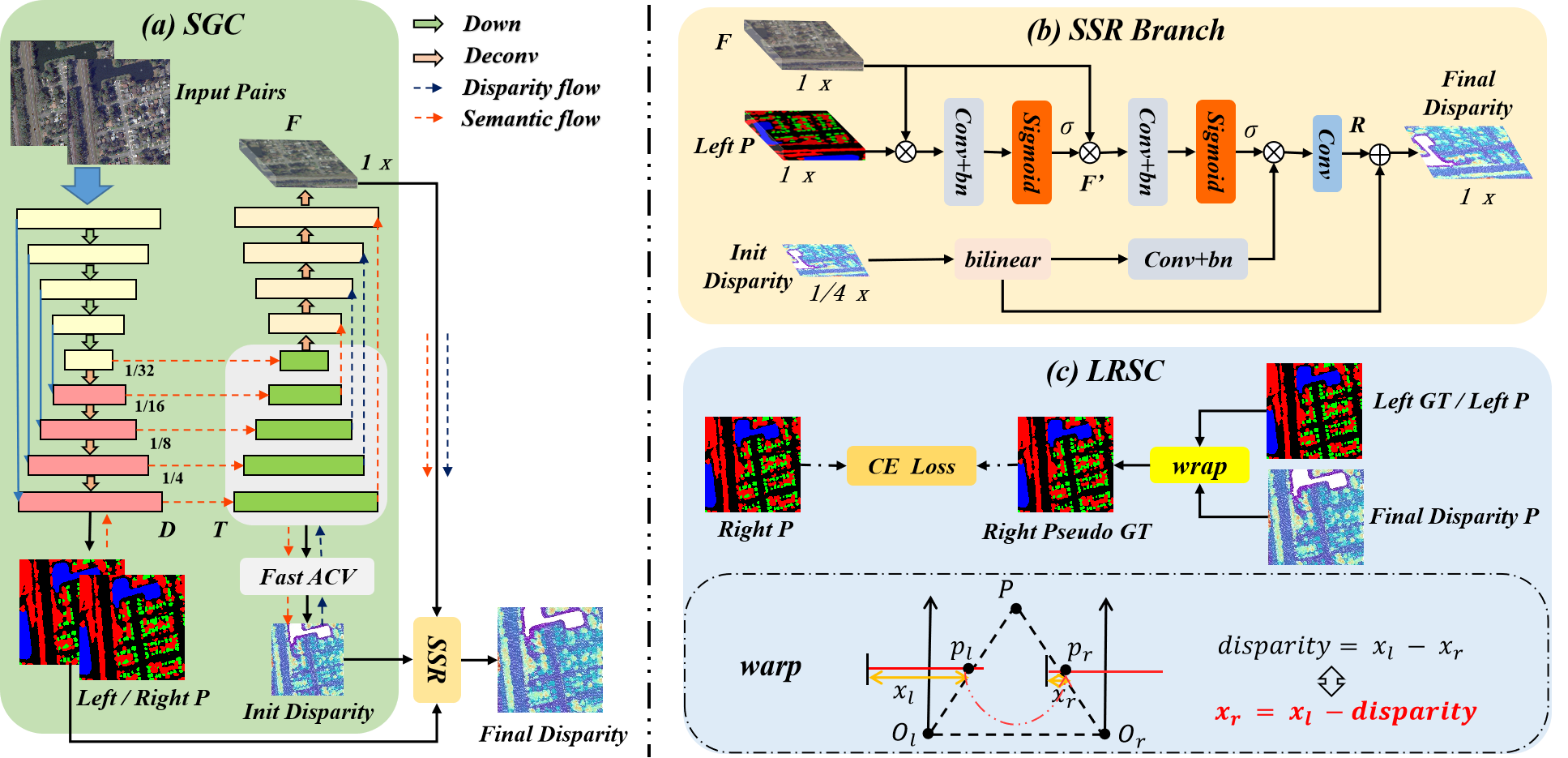}
\caption{\label{fig:2}
An overview of the SemStereo. It involves (a) a Semantic-Guided Cascade (SGC) structure for generating segmentation and initial disparity maps, (b) a Semantic Selective Refinement (SSR) branch refines the initial disparity under the guidance of semantic information, and (c) a Left-Right Semantic Consistency (LRSC) supervision. P: Prediction, GT: Ground Truth.
}
\end{figure*}

%% file: sec/03_method.tex
\section{Method}
\label{sec:method}

In this section, we provide a comprehensive description of our SemStereo, as illustrated in \Cref{fig:2}.
It involves a Semantic-Guided Cascade (SGC) structure for the generation of an initial disparity, a Semantic Selective Refinement (SSR) branch that refines the disparity under the guidance of semantic segmentation maps, and the Left-Right Semantic Consistency (LRSC) constraints on both views.
% }
%, comprising a Semantic-Guided Cascade structure, a Semantic Selective Refinement branch, and Left-Right Semantic Consistency supervision. 

\subsection{Semantic-Guided Cascade Structure}

%\quad Semantic Segmentation guided framework leverages semantic segmentation as prior knowledge to improve object boundary accuracy and maintain semantic consistency in the stereo-matching task.
%In our Semantic-Guided Cascade framework, we leverage the U-shape structure to integrate the feature extractor of stereo matching and the semantic segmentation encoder-decoder network, creating a Shared Siamese U-shape Feature Extractor.
%to achieve task consistency between the two heterogeneous tasks

%To fully leverage the task consistency, 
%Our Semantic-Guided Cascade feature extractor introduces Shared Siamese U-shaped networks that unify the semantic Encoder-Decoder networks and the feature extractor of stereo matching. %Building upon this, we construct a Semantic-Guided Cascaded Cost Volume for stereo matching.
% \textcolor{blue}{
The Semantic-Guided Cascade structure (\Cref{fig:2} (a)) incorporates a U-shaped network for extracting deep shared features for both tasks, along with a Fast-ACV for the disparity computation. 
This design facilitates deeper feature sharing between the two tasks, thereby strengthening the implicit influence of semantic cues on stereo matching.
% }
% Our Semantic-Guided Cascade structure (Fig. \ref{fig:2} (a)) incorporates a Shared Siamese U-shaped feature extractor, unifying the semantic encoder-decoder network with the stereo matching feature extractor. 
% This integration is designed to enable the two tasks to share deeper features, 
% thereby enhancing the implicit constraints of semantic clues on stereo matching.
% consistency at the feature level.

% \subsubsection{Shared Siamese U-shape Feature Extractor}
% \subsubsection{Shared U-shape Feature Extractor}
%Inspired by the development of semantic segmentation networks~\cite{long2015fully, chen2017rethinking, zhao2017pyramid, xie2021segformer}, we notice that the cross-scale and contextual awareness capabilities of feature extraction in stereo matching are often overlooked, and the lack of cross-scale ability may result in the loss of details, while the lack of contextual awareness may lead to poor performance in textureless regions. Meanwhile, the U-shape network architecture for semantic segmentation continues to be widely researched and applied, demonstrating strong robustness. 
\textbf{Shared U-shape Feature Extractor.}
% \textcolor{blue}{
The U-shaped feature extractor generates shared features for both semantic segmentation and disparity estimation, taking as input an image pair $I^{l}, I^r \in \mathbb{R}^{3\times H \times W}$.
% }
% \textcolor{blue}{
Given the high-resolution input of remote sensing images and the need for efficient extraction of robust features with global receptive fields, we employ MobileViTv2~\cite{mehta2021mobilevit, mehta2022separable} as the encoder. This is followed by a decoder with skip connections and transposed convolutions, which progressively derives features at multiple scales ${D^{l}_i, D^r_i \in \mathbb{R}^{C'_i \times \frac{H}{i} \times \frac{W}{i}}}$ ($i=2, 4, 8, 16, 32$).
% }

\textbf{Semantic Segmentation.}
% \textcolor{blue}{
Attached to the deepest feature volume $D^l_{2}, D^r_{2}$ is a simple module for semantic segmentation.
It involves a convolution, upsampling, and softmax layer, resulting in the pixel-wise segmentation heat map denoted as $P^l, P^r \in \mathbb{R}^{N \times H \times W}$ ($N$ for the number of semantic classes, the same below).
% }

% Considering the high-resolution input of remote sensing images, and to efficiently extract robust features with global receptive fields, we adopt MobileViTv2 \cite{mehta2021mobilevit, mehta2022separable}
% , a combination of MobilenetV2\cite{sandler2018mobilenetv2} blocks and mobile vision transformers, 
% for the shared siamese encoder. 
% Given an image pair 
% $I_{l(r)} \in \mathbb{R}^{3\times H \times W}$, 
% we apply a stride $3 \times 3$ standard convolution, followed by MobileViTv2 blocks, to obtain feature map pairs at different levels denoted as ${E_{l(r), i} \in \mathbb{R}^{C_i \times \frac{H}{i} \times \frac{W}{i}}}$ ($i=2, 4, 8, 16, 32$). 
% Then we utilize a siamese decoder that includes skip connections and transposed convolutions to progressively upsample $E_{l(r), i}$ to the original resolution denoted as ${D_{l(r), i} \in \mathbb{R}^{C'_i \times \frac{H}{i} \times \frac{W}{i}}}$ ($i=2, 4, 8, 16, 32$).

% \subsubsection{Semantic-Guided Cascade Cost Volume} 
% \textbf{Semantic-Guided Cascade Cost Volume for Disparity Computation.}
\textbf{Cost Volume for Disparity Computation.}
% Unlike previous multi-task learning models~\cite{yang2018segstereo, zhang2019dispsegnet, dovesi2020real}, which share only the shallow parts between two tasks (\Cref{fig:framework_comparison}~(a)), 
% \textcolor{blue}{
Aiming at strengthening the semantic influence on stereo matching, 
we adopt a cascaded structure (\Cref{fig:framework_comparison}) that utilizes deep shared features $D^l_i$ and $D^r_i$, enriched with semantic information, for initial disparity map generation.
This approach contrasts with earlier methods that relied on shallow features, which resulted in weaker connections between tasks~\cite{yang2018segstereo, zhang2019dispsegnet, dovesi2020real}.
% }
% \textcolor{blue}{
Specifically, the feature volumes $D^l_i, D^r_i$ are firstly fed into 
a series of $1\times1$ siamese convolutions to bridge the gap between tasks and halve the number of channels to ensure an equitable comparison with the baseline \cite{acv, xu2023accurate}, yielding feature volumes $T^l_i, T^r_i \in \mathbb{R}^{C''_i \times \frac{H}{i} \times \frac{W}{i}}$, where $C''_i = \frac{C'_i}{2}$. 
Then we construct the attention concatenation cost volume $V \in \mathbb{R}^{C''' \times \frac{D_{max}}{2} \times \frac{H}{4} \times \frac{W}{4}}$ as Fast-ACV~\cite{acv, xu2023accurate} with a minor adaptation for remote sensing that the disparity spans from negative ($-D_{max}$) to positive values ($D_{max}-1$).
The initial disparity map $d_{init} \in \mathbb{R}^{1 \times \frac{H}{4} \times \frac{W}{4}}$ is derived after regulating the cost volume $V$ as Fast-ACV~\cite{acv, xu2023accurate}.
% , while we construct volume  spanning from negative disparities to positive by shifting right features from $-D_{max}$ to $D_{max}-1$, as \cite{he2021dsm, he2022hmsm}.
% }
% Specifically, we attach a semantic segmentation classifier head after the decoder of shared siamese U-shape networks for semantic supervision to get the semantic segmentation prediction results of the source (left) image, denoted as $L^{p} \in \mathbb{R}^{N \times H \times W}$ (N for the number of semantic classes).

\textbf{Feature volume $F$ for SSR.}
In addition to the segmentation map and the initial disparity, the third output of SGC is a feature volume containing both semantic and disparity information, as the disparity flow and semantic flow depicted in ~\Cref{fig:2} (a), which is then fed to SSR for further disparity refinement. 
%Given features $D_i$ from the shared U-shape networks, each scale $i$ is followed by a $1\times1$ siamese convolution to bridge the gap between tasks and halve the number of channels to ensure an equitable comparison with the baseline \cite{acv, xu2023accurate}, yielding feature pairs denoted as $T_i$ with dimensions of $C''_i \times \frac{H}{i} \times \frac{W}{i}$ ($i=2, 4, 8, 16, 32$ and $C''_i = \frac{C'_i}{2}$).
It generates a comprehensive feature volume $F \in \mathbb{R}^{N \times H \times W}$ by progressively upsampling and concatenating the features $T_i$.

\input{tables/2}

\subsection{Semantic Selective Refinement Branch}

To explicitly leverage intra-class disparity consistency in remote sensing scenes (\Cref{fig:disparity_category}), we introduce the Semantic-Selective Refinement (SSR) branch. This branch uses a channel attention mechanism to selectively learn disparity residual errors from the semantic prediction, which are then applied to refine the initial disparity map.
% }

% \textcolor{red}{
% Considering that semantic predictions may introduce new ambiguities, especially in the absence of explicit semantic supervision or during the early stages of training, we employ the feature volume $F$ from adapters to learn the semantic information selectively, instead of directly fusing semantic predictions with disparities.
% }

% \textcolor{blue}{
As shown in~\Cref{fig:2} (b), we first compute the inner production of the feature volume $F$ and the predicted semantic map $P^l \in \mathbb{R}^{N \times H \times W}$. Notably, we preserve the multi-channel outputs from the semantic segmentation, where each channel represents the probability map for a specific class, thereby maintaining the confidence level of each pixel belonging to a certain class. Next, we compute their correlation score using a $1\times1$ point convolution followed by Batch Normalization and a Sigmoid activation function to generate a weight map $\sigma(F\cdot P^l)$. This weight map is then applied to filter the feature volume $F$, yielding a new feature volume $F' \in \mathbb{R}^{N \times H \times W}$ as follows:
% }

% As is shown in~\Cref{fig:2} (b), we first multiply the feature volume from the adapters $F$ with the predicted semantic map $L^p \in \mathbb{R}^{N \times H \times W}$. 
% Then we compute their correlation score by utilizing $1\times1$ point convolution followed by Batch Normalization and Sigmoid activation function. 
% It is worth noting that we retain the multi-channel results of semantic segmentation, where each channel represents the probability map for that class, preserving the confidence in each spatial pixel being a certain class. 
% Subsequently, we re-weight the feature volume $F$ to obtain semantic probabilistic features $F' \in \mathbb{R}^{N \times H \times W}$ through the computed score as:

%Each channel of the semantic prediction result represents the probability map for that category, which when multiplied by the features $I$, yields the features selected for each semantic category.

\begin{equation}
{F'} = \sigma(F\cdot P^l)\cdot F,
\end{equation}

\noindent where $\sigma$ denotes $1\times1$ CNN followed by Batch Normalization and Sigmoid.

% \input{figs/4}

%We also obtain the up-sampled initial disparity map $d_{init}' \in \mathbb{R}^{1 \times H \times W}$ utilizing bilinear upsampling. Meanwhile, we transform the selective semantic residuals $I'$ into a probability measure $P' \in \mathbb{R}^{N \times H \times W}$. Then we add the semantic residuals with $d_{init}'$ to get the final disparity map $d_{final}$:

Furthermore, we apply the new feature volume $F'$ to generate a new weight map $\sigma(F')$ to filter the initial disparity map.
Specifically, we perform bilinear upsampling on $d_{init}$ to obtain $d_{init}' \in \mathbb{R}^{1 \times H \times W}$. 
Subsequently, we normalize $d_{init}'$ and use a $3\times3$ convolution to expand the channels to the number of classes, denoted as $d_{init}'' \in \mathbb{R}^{N \times H \times W}$. 
We multiply $d_{init}''$ with the new weight map $\sigma(F')$.
% to obtain residuals. 
Finally, we utilize a $1 \times 1$ CNN to obtain single channel residuals $R$ and refine the final disparity map $d_{final}$ by adding with $d_{init}''
$:%we concatenate the residuals and the upsampling disparity in the channel dimension and add a $1 \times 1$ CNN to learn the weights of them, obtaining the final disparity result $d_{final}$.

\begin{equation}
R = Conv(\sigma(F')\cdot d_{init}''),
\end{equation}

% \begin{equation}
% {d_{final}} = Conv(d_{init}' \  \textcircled{c} \ residuals)
% \end{equation}
\begin{equation}
{d_{final}} = R + d_{init}' .
\end{equation}

%\noindent where $\sigma$ denotes $1\times1$ CNN followed by BN and Sigmoid.%, and N denotes feature channels and is equal to the number of semantic classes.

\subsection{Left-Right Semantic Consistency Supervision}
%In semantic segmentation tasks, training only the features from the left image for prediction may cause the network's weights to be biased towards the representation of the left image. But in stereo disparity estimation, it is necessary to analyze the relative relationship between the left and right images, and the representation of the right image is equally important. 
%However, in general scenarios, obtaining manual labels for the right image is often expensive, and typically only the semantic labels for the left image are available. To tackle this problem, we note that there exists a deterministic mapping relationship between disparity and the semantic labels of the left and right images. 

%\input{figs/12}
%To address this issue, an effective approach is to increase supervision of the right image. and to increase the diversity of the sample for semantic supervision
% \textcolor{blue}{
We further impose a semantic consistency constraint between both views after warping the semantic segmentation map (or feature map in the absence of pixel-wise annotations) of the left view to the right view based on the refined disparity map. 
% }
% To further leverage inter-task consistency across views, we propose another explicit constraint, named Left-Right Semantic Consistency supervision.
%We warp the left semantic labels according to their corresponding disparities and use the warped right semantic labels as pseudo-labels to supervise the semantic networks of the right image. This helps constrain the semantic consistency between the left and right features and promotes inter-task consistency between semantic and disparity tasks.

%Because the definition of disparity is the pixel difference in the x-axis direction between corresponding points in the left and right images, it is possible to infer one image from the disparity map and one of the images.
Disparity measures the horizontal pixel difference between corresponding points in left ($p_l$) and right ($p_r$) images:
\begin{equation}
disparity = x_l -x_r,
\end{equation}
where $x_l$ and $x_r$ are the horizontal coordinates of $p_l$ and $p_r$. 
Therefore, we leverage this relationship by warping the semantic labels of the left view $GT^l$ to the right using the corresponding disparity, thus obtaining pseudo-semantic labels $GT^r$ for the right semantic supervision (See \Cref{fig:2} (c)).
% Considering the discontinuity of disparity labels, 
%As shown in Fig.~\ref{fig:intro2}, in large-scale urban scenes, directly using disparity labels for warping introduces holes and inconsistencies. To mitigate the impact of holes and the inconsistent between the two tasks, 
% we employ the disparity predictions $d_{final}$ rather than the disparity labels $d^{gt}$ to obtain the right semantic pseudo-labels $R^{gt}$. 
% \textcolor{blue}{
Additionally, given the high cost of semantic annotations, we account for scenarios where such annotations may be unavailable. In these cases, simply replacing $GT^l$ with $P^l$ enables our method to operate in a self-supervised manner.
% }

% Additionally, due to the high cost of semantic annotations, we take into account scenarios where semantic annotations may not be available. 
% In such a case, it is only necessary to replace $L^{gt}$ with $L^{p}$, thus converting the semi-supervised method into a self-supervised method.

% \textcolor{blue}{
To achieve that goal, we minimize the discrepancy between the semantic maps after warping using the Cross-Entropy (CE) loss as an auxiliary loss $\mathcal{L}_{LRSC}$:
% }
 % We introduce a right semantic segmentation prediction head to generate the right semantic predictions $R^{p}$. These predictions are aligned with the warped right semantic pseudo-labels $R^{gt}$ using the Cross-Entropy (CE) loss as an auxiliary semi-/self-supervised loss $\mathcal{L}_{LRSC}$: %This results in consistent geometric structures among the outputs of different tasks.
\begin{equation}
% R^{gt} = warp(L^{gt} or L^{p}, d_{final}),
% \[
R^{gt} =
\begin{cases}
\text{warp}(GT^l, d_{final}), & \text{if } GT^l \text{ is available} \\
\text{warp}(P^l, d_{final}), & \text{if } GT^l \text{ is not available}
\end{cases}
% \]
\end{equation}
\begin{equation}
\mathcal{L}_{LRSC} = \mathcal{L}_{CE}(P^r, GT^r).
\end{equation}

%This alignment helps ensure that the right semantic predictions are consistent with the pseudo-labels, further improving the semantic inter-task consistency. %between the left and right images in the framework.

\subsection{Loss Function}

\quad For the semantic segmentation task, we combine Dice loss and CE loss to leverage the strengths of both:

\begin{equation}
\mathcal{L}_{Seg} =  \mathcal{L}_{CE}(L^{p}, L^{gt}) + \mathcal{L}_{Dice}(L^{p}, L^{gt}).
\end{equation}

% \noindent where $\alpha$ is a hyperparameter that controls the relative weights of the two losses.

For the stereo matching task, Smooth $L_1$ loss is used:
% \cite{psm}
\begin{equation}
\mathcal{L}_{Disp} = \sum_{i}^{n}\lambda_i \mathcal Smooth_{L1}(d_{i} - d^{gt}),
\end{equation}
where $d_{i}$ are the predicted disparity map of different stages, and $d^{gt}$ is the disparity ground truth, $\lambda_{i}$ are hyperparameters that control the relative weights of the different stages.

For the entire multi-task model, our joint loss is given by,

\begin{equation}
\mathcal{L} =  \mathcal{L}_{Disp} + \alpha \mathcal{L}_{Seg} + \beta \mathcal{L}_{LRSC},
\end{equation} 

\noindent where $\alpha$ and $\beta$ are hyperparameters to control the relative weights of losses.% 

%% file: tables/2.tex
\begin{table*}[!t]
\caption{\label{tab:2}Ablation quantitative evaluation of Semantic Guided-Cascade (SGC) framework, Semantic Selective Refinement (SSR) branch, and Left-Right Semantic 
Consistency (LRSC) supervision on US3D Jacksonville test set. * : Without explicit semantic label supervision. \textbf{Bold}: Best. } %~\cite{US3D, IADF_GRSM_201903}
\fontsize{9}{11}\selectfont
\centering
% \vspace{-0.8cm}   %调整图片与上文的垂直距离  
% \setlength{\abovecaptionskip}{0.cm} %调整标题上方的距离   
% \setlength{\abovecaptionskip}{0.cm} %调整标题下方的距离 
\begin{tabular}{cccccccccccccccccc}
\toprule
& \multirow{2}{*}{Model} & \multirow{2}{*}{SGC} & \multirow{2}{*}{SSR} & \multirow{2}{*}{LRSC} & \multicolumn{2}{c}{Stereo Matching} && \multicolumn{2}{c}{Semantic} \\
\cline{6-7} \cline{9-10}
& &  &  &  &EPE(Pixel)$\downarrow$ & D1(\%)$\downarrow$ && mIOU(\%)$\uparrow$ & PA(\%)$\uparrow$\\
  % &  &  &  &&&  &  &  &\\
\midrule
1 & Baseline & &  && 1.2087 & 7.28 && 75.84 & 93.65\\
 %S3G &\checkmark &&  & 1.0231 & 5.49 & 78.79 & 93.68\\
2 & SGC-Net & \checkmark &  && 0.9995 & 4.98 && 75.74 & 93.70\\
% SR-Net & & \checkmark &  & 1.0931 & 5.89 & 12.12 & 33.31 && 76.59 & 93.68\\
% SD-Net & &  & \checkmark & 1.2071 & 7.25 & 13.21 & 36.03 && 76.49 & 93.86\\
3 &SGC-SSR-Net & \checkmark & \checkmark && 0.9702 & 4.76 && 76.85 & 93.83\\

% SG-SD-Net & \checkmark &  & \checkmark & 0.9791 & 4.90 & 9.58 & 29.57 && \textbf{77.38} & 94.08\\
% SD-SR-Net &  & \checkmark & \checkmark & 1.0738 & 5.78 & 11.08 & 32.81 && 76.04 & 94.02\\
4 &SemStereo & \checkmark & \checkmark & \checkmark & \textbf{0.9582} & \textbf{4.58} && \textbf{77.02} & \textbf{94.13}\\
\midrule
5 &Baseline* &  &  && 1.2260 & 7.47 && - & -\\
6 &SGC-Net* & \checkmark &  && 1.0499 & 5.61  &&- & -\\
7 &SGC-SSR-Net* & \checkmark & \checkmark && 1.0164 & 5.34 && - & -\\
8 &SemStereo* & \checkmark & \checkmark & \checkmark & \textbf{0.9956} & \textbf{5.00} && - & -\\
\bottomrule
\end{tabular}
\end{table*}

%SG-3 & &  & \checkmark &  &  & \checkmark & 1.0167 & 5.41 & - & - & 173\\
%SSG-1 & \checkmark &  &  & \checkmark &  & \checkmark & 1.0171 & 5.37 & 76.07 & 93.23 & 176\\
%SSG-2 &  & \checkmark &  & \checkmark &  & \checkmark & 1.0154 & 5.20 & 74.92 & 93.32 & 179\\
%SSG-3 &  &  & \checkmark & \checkmark &  & \checkmark & 0.9995 & 4.98 & 76.94 & 93.70 & 175\\
%SR-SG &  &  & \checkmark &  & \checkmark & \checkmark &  1.0164 & 5.34 & - & - & 180\\
%SR-P &  &  & \checkmark &  & \checkmark &  & 1.0672 & 5.96 & - & - & \textbf{141}\\
%SSR-P &  &  & \checkmark & \checkmark & \checkmark &  & 1.0231 & 5.49 & 77.79 & 93.68 & 172\\

%% file: sec/04_experiment.tex
\input{tables/4}
\input{tables/gen}

\section{Experiments}
\label{sec:experiments}
\input{figs/5}

\subsection{Datasets}

%We conduct experiments on US3D (Track 2)~\cite{IADF_GRSM_201903, le20192019, US3D} and WHU~\cite{liu2020novel}.%, representing satellite scenes and aerial scenes respectively. %In addition, we combined KITTI 2015 \cite{menze2015object} for extended validation to compare with other semantic stereo methods.

%We conduct experiments on remote sensing urban scenes dataset US3D (Track 2)~\cite{IADF_GRSM_201903, le20192019, US3D}. Additionally, we incorporate KITTI \cite{menze2015object} for extended validation. %Scene Flow \cite{mayer2016large} 
%~\cite{US3D, IADF_GRSM_201903}

% \textbf{US3D}~\cite{IADF_GRSM_201903, le20192019, US3D} is a satellite dataset, containing 2139 pairs of stereo images of Jacksonville and 2153 pairs of Omaha with corresponding semantic labels. 
% We randomly select Jacksonville 1500, 139, and 500 pairs as training, validation, and test sets respectively, and Omaha as generalization verification. 
% \textcolor{blue}{
\textbf{US3D}~\cite{US3D} 
contains
% is a satellite dataset comprising 
2,139 pairs of satellite stereo images from Jacksonville and 2,153 from Omaha, each with corresponding semantic labels. 
We randomly select 1,500 pairs from Jacksonville for training, 139 for validation, and 500 for testing, and use Omaha for generalization verification.
% }

%The pairs are geographically non-overlapping tiles rectified to a size of 1024×1024.

% \noindent\textbf{WHU}~\cite{liu2020novel} is an aerial dataset generated from thousands of real aerial images, covering 6.7×2.2 km$^2$ over Meitan County, China, with a ground resolution of about 0.1 m, but no corresponding semantic labels.
% \textcolor{blue}{
\textbf{WHU}~\cite{liu2020novel} is an aerial dataset from 8,316 real aerial images in the training set and 2,618 in the test set, covering an area of $6.7\times2.2\ km^2$ over Meitan County, China, with a ground resolution of approximately 0.1 meters. However, it lacks corresponding semantic labels.
% }
\input{tables/semantic}
\input{figs/6}

%The coverage area contains dense high-rise buildings, sparse factories, forested mountains, and some bare ground and rivers.
%A random seed is used to randomly select the training pairs (see Table ~\ref{tab:1}).

%\input{tables/1}

%\textbf{KITTI 2015} is a dataset for real-world driving scenes, containing 200 training pairs and 200 testing pairs, each accompanied by corresponding per-pixel semantic labels. % that follow the same specification as Cityscapes \cite{cordts2016cityscapes}.\cite{menze2015object} \cite{menze2015object}

\input{figs/13}

\subsection{Implementation Details}

% We implement our SemStereo with PyTorch and conduct our experiments utilizing NVIDIA A40 GPUs. 
% The loss hyperparameters are set as $\lambda_0=1$, $\lambda_1=0.6$, $\lambda_2=0.5$, $\lambda_3=0.3$, $\alpha, \beta=1$.
% \textcolor{blue}{
We implement SemStereo using PyTorch and conduct our experiments on two NVIDIA A40 GPUs.
The hyperparameters for the loss function are set as follows: $\lambda_0=1$, $\lambda_1=0.6$, $\lambda_2=0.5$, $\lambda_3=0.3$, and $\alpha, \beta=1$.
% }
%We also halve the number of channels in the feature extractor, remove the upsampling part for 1/2-scale features, and replace the fast attention concatenation volume~\cite{acv, xu2023accurate} with group-wise correlation volumes~\cite{gwc} in stereo network, constructing our lightweight model version called SemStereo-s. 
%As US3D~\cite{IADF_GRSM_201903, le20192019, US3D} includes reasonable negative disparities in its disparity range, we reconstruct state-of-the-art models to ensure their applicability to large-scale urban scenes. 
% \textcolor{blue}{
We compare our approach with a range of state-of-the-art methods from both the computer vision and remote sensing communities.
% }
% We introduce state-of-the-art methods from driving scenarios into remote sensing scenarios. %For all training, we employ Adam optimizer with $\beta_1 = 0.9$ and $\beta_2 = 0.999$. 
% For fairness, we unify the configuration parameters, set the optimizer to Adam optimizer with $\beta_1 = 0.9$ and $\beta_2 = 0.999$, the batch size to 4, and keep the original resolution without any augmentation techniques. 
% Our training epoch number for a single stage is set to $48$, and the initial learning rate ($lr_0$) is set to $0.001$, decaying by half after epochs $12, 22, 30, 38$, and $44$. 
% The range of disparity varies across different datasets; referring to previous work, US3D is set to $[-64, 64)$ and WHU is set to $[0, 128)$.
% \textcolor{blue}{
For fairness, we standardize the configuration parameters: the optimizer is set to Adam with $\beta_1 = 0.9$ and $\beta_2 = 0.999$, the batch size is 4, and we use the original resolution without any augmentation techniques. We train each stage for 48 epochs, starting with an initial learning rate ($lr_0$) of 0.001, which decays by half after epochs 12, 22, 30, 38, and 44. The disparity range varies by dataset: US3D is set to $[-64, 64)$ and WHU is set to $[0, 128)$, following the settings used in previous work~\cite{he2021dsm}.
% }
%., and KITTI is set to $[0, 196)$.% a disparity range of $[-64, 64)$, 
%We train the model and conduct validation in Jacksonville. In addition, we conduct transfer learning experiments with a limited dataset in Omaha to evaluate the generalization capacity.
%as Fig.~\ref{fig:10}

%\input{tables/ab2}
\subsection{Ablation Study}

% \subsubsection{Semantic Stereo Scenes}
% We perform ablation experiments to compare our Semantic-Guided Cascade (SGC) framework with common parallelized frameworks (\Cref{fig:framework_comparison}), Semantic Selective Refinement (SSR) with the common bilinear upsampling method~\cite{psm, gwc}, and with or without Left-Right Semantic Consistency (LRSC).
% \textcolor{blue}{
We conduct ablation experiments to assess the effectiveness of our proposed Semantic-Guided Cascade (SGC) structure by comparing it with standard parallelized frameworks~\cite{zhang2019dispsegnet, dovesi2020real, liao2023s} (\Cref{fig:framework_comparison}), the Semantic Selective Refinement (SSR) by replacing it with a common bilinear upsampling method~\cite{psm, gwc}, and the Left-Right Semantic Consistency (LRSC) by removing it.
% }
%As shown in Table~\ref{tab:2}, experimental results show that SGC, SSR, and LRSC are of great help in improving the accuracy of semantic segmentation and stereo matching, especially when the three are integrated. 
%As shown in Table~\ref{tab:2}, compared with the parallel structure, SGC has an improvement of 31.6\% in the D1 metric and 17.3\% in the EPE metric. With the introduction of SSR, the D1 metric further increases by 4.4\% and the EPE metric by 2.9\%. The introduction of LRSC further leads to an improvement of 3.8\% in D1 and 1.2\% in EPE. Our method remains effective even without explicit semantic label supervision. The introduction of SGC brings an improvement of 24.9\% in D1 and 14.4\% in EPE. The introduction of SSR results in a further increase of 4.8\% in D1 and 3.2\% in EPE, and the introduction of LRSC leads to a further improvement of 6.4\% in D1 and 2.0\% in EPE. By comparison, when semantic supervision is introduced, SGC enhances D1 by 11.2\% and EPE by 4.8\%, SSR improves D1 by 10.9\% and EPE by 4.5\%, and LRSC boosts D1 by 8.4\% and EPE by 3.7\%.

% \textcolor{blue}{
As shown in Table~\ref{tab:2}, the SGC structure improves the baseline method by 31.6\% in the D1 metric and 17.3\% in the EPE metric (Line 2 vs. Line 1), demonstrating its effectiveness.
Comparing Line 2 with Line 6, we observe that incorporating semantic supervision results in an 11.2\% improvement in D1 and a 4.8\% improvement in EPE, underscoring the critical role of semantic information in enhancing stereo matching.
The introduction of SSR further enhances the D1 metric by 4.4\% and the EPE by 2.9\% (Line 2 vs. Line 3).
% }

% \textcolor{blue}{
The introduction of LRSC improves the D1 metric by 3.8\% and the EPE metric by 1.2\% (Line 3 vs. Line 4). 
In the absence of explicit semantic annotations, LRSC still enhances D1 by 6.4\% and EPE by 2.0\% (Line 7 vs. Line 8).
Additionally, incorporating semantic supervision results in an 8.4\% improvement in D1 and a 3.7\% improvement in EPE (Line 4 vs. Line 8). 
This demonstrates that LRSC effectively enhances performance by enforcing semantic consistency across views, both with and without explicit semantic supervision, and achieves even better results when semantic annotations are provided.

\subsection{Comparisons with State-of-the-art}
% \subsubsection{Performance of Stereo Matching}
\textbf{Stereo Matching}
% \textcolor{blue}{
We compare our SemStereo with the state-of-the-art stereo methods on US3D~\cite{IADF_GRSM_201903, le20192019, US3D} and WHU~\cite{liu2020novel},
as presented in Table~\ref{tab: 4}.
% }
%Considering the requirement of multiple stages training such as ACVNet \cite{acv}, we conduct two separate experiments to ensure fairness. In the first experiment, a single-stage training approach is employed, while in the second, models requiring multi-stage training are trained using respective multi-stage methods, and other models are fine-tuned for an additional cycle of the same epochs.

% \textcolor{blue}{
When semantic labels are unavailable, our downgraded model, SemStereo*, still surpasses state-of-the-art methods on both US3D and WHU datasets. 
With the introduction of semantic labels during training, our full model, SemStereo, achieves an additional 14.6\% improvement in the D1 metric, representing a more substantial enhancement compared to previous methods. SemStereo outperforms IGEV-Stereo~\cite{igev} by 37.7\%, Fast-ACVNet~\cite{acv,xu2023accurate} by 35.4\%, HMSMNet~\cite{he2022hmsm} by 42.35\%, GwcNet~\cite{gwc} by 34.76\%, and PSMNet~\cite{psm} by 33.6\% on the D1 metric.
%, and the small model SemStereo-s improves by 9.05\%
% }
%Taking an example, our small full model SemStereo-s improves by 14\% on Runtime, 16\% on D1 metric, and 8.6\% on EPE metric compared to Fast-ACVNet. 

%To make it more intuitive, we plot the performance comparison of speed and time in Fig.~\ref{fig:1}.

% To verify the generalization ability across cities, we also perform transfer learning on limited data of Omaha.
%We randomly select 50 and 500 pairs from Omaha for fine-tuning of 12 and 48 epochs respectively, and 1500 pairs are used as verification and compare the generalization of models. 
%Our SemStereo-s is observed to have better speed and better generalization ability than other fast models, such as StereoNet~\cite{li2018stereo} and Fast-ACVNet~\cite{acv, xu2023accurate}, while 
%Our SemStereo is observed to have the best generalization accuracy, as is shown in Table~\ref{tab: 5}.% compared to all previous models.

% \textcolor{blue}{
To evaluate generalization across cities, we assess zero-shot capabilities and conduct transfer learning with limited data from Omaha. 
We randomly select 50 and 500 pairs from Omaha for fine-tuning over 12 and 48 epochs, respectively, using 1500 pairs for validation. 
As shown in Table~\ref{tab: 5}, our model achieves state-of-the-art performance with an EPE of 1.4996 in zero-shot scenarios. 
As the amount of fine-tuning data increases, our model’s performance improves. 
When fine-tuning is performed with 500 pairs, our model’s performance on Omaha closely matches the results from the original city (Jacksonville), with D1 scores of 4.54\% vs. 4.58\%. 
Additionally, our model demonstrates a clear advantage over other models, improving D1 by 9.9\% and EPE by 2.5\%.

% To verify the generalization ability across cities, we compare the zero-shot capabilities and conduct transfer learning with limited data on Omaha. 
% We randomly select 50 and 500 pairs from Omaha for fine-tuning over 12 and 48 epochs, respectively, with 1500 pairs used for validation. 
% The experimental results presented in Table~\ref{tab: 5} reveal that
% our model achieves SOTA performance in terms of EPE (1.4996) in the zero-shot scenes.
% % The average performance in D1 is speculated to be due to certain mechanisms within our baseline, Fast-ACNet \cite{acv,xu2023accurate}, as the generalization performance of Fast-ACVNet is relatively poor, and the significant improvement of our model over it also suggests this. 
% As fine-tuning gradually expands, the advantages of our model also grow. 
% When the fine-tuning samples increase to 500 pairs, our model's performance on the Omaha is comparable to the original city's (Jacksonville) results (4.54\% vs 4.58\% in the D1 metric), with a clear advantage over other models (improves 9.9\% in the D1 and 2.5\% in the EPE).
% }

% to simulate the generalization ability from one city to another
% \textcolor{blue}{
A qualitative comparison of stereo matching results on the US3D test set is shown in \Cref{fig:5} and \Cref{fig:6}.
SemStereo exhibits clearer boundaries and more detailed features, owing to enhanced feature interactions between tasks and the incorporation of explicit inter-task constraints.
% }

% A qualitative comparison of stereo matching on the US3D test set is shown in \Cref{fig:5} and \Cref{fig:6}. Our SemStereo demonstrates clearer boundaries and richer details, thanks to the increased feature interaction between tasks and the introduction of explicit inter-task constraints.

%\input{tables/3}

\subsubsection{Semantic Segmentation} 
% \textcolor{blue}{\textbf{.}}
% \textcolor{blue}{
As shown in Table~\ref{tab: 6}, we compare our model with established semantic segmentation networks, including FCN~\cite{long2015fully}, UNet~\cite{ronneberger2015u}, DeepLabV3~\cite{chen2017rethinking}, PSPNet~\cite{zhao2017pyramid}, and SegFormer~\cite{xie2021segformer}.
SemStereo achieves state-of-the-art results, outperforming SegFormer by 14.45\% in mIoU and PSPNet by 9.96\% in mIoU. 
Additionally, our model shows significant advantages over other semantic stereo methods, such as S$^2$Net and S$^3$Net~\cite{liao2023s,yang2024s3net}. The inclusion of stereo matching supervision enhances semantic accuracy by 9.45\% mIoU in the full SemStereo. 
Qualitative results in \Cref{fig:12} demonstrate that SemStereo delivers clearer boundaries on dense building clusters and large objects while preserving more details, attributable to the enhanced disparity label supervision.

%% file: tables/4.tex
\begin{table*}[!t]
\caption{\label{tab: 4}Quantitative comparison of stereo matching between our SemStereo and state-of-the-art models on US3D and WHU test set. \dag
: The results are obtained from the official declaration. * : Without explicit semantic label supervision. \textbf{Bold}: Best. }
\fontsize{9}{11}\selectfont
\centering
\begin{tabular}{ccccccccccccc}
\toprule
\multirow{2}{*}{Method} &  \multicolumn{2}{c}{US3D} && \multicolumn{2}{c}{WHU} \\
\cline{2-3} \cline{5-6}
 % &  EPE & D1 & Thres2 & Thres1 && EPE & D1 & Thres2 & Thres1 & \multirow{2}{*}{(ms)$\downarrow$}\\ % 
 & EPE(Pixel)$\downarrow$ & D1(\%)$\downarrow$ && EPE(Pixel)$\downarrow$ & D1(\%)$\downarrow$\\
\midrule
StereoNet \cite{khamis2018stereonet} & 1.6053 & 12.13 && 0.3881 & 1.413\\
S$^2$Net\dag
 \cite{liao2023s} & 1.439 & 10.05 && - & -\\
S$^3$Net\dag
 \cite{yang2024s3net} & 1.403 & 9.58 && - & -\\
DSMNet\dag \cite{he2021dsm} & 1.2776 & 7.94 && - & -\\
HMSMNet \cite{he2022hmsm} & 1.2338 & 7.91 && 0.2745 & 0.904\\
GwcNet \cite{gwc} & 1.2120 & 6.99 && 0.2549 & 0.862\\
PSMNet \cite{psm} & 1.1770 & 6.87 && 0.2432 & 0.814\\
IGEV-Stereo \cite{igev} & 1.2051 & 7.32 && - & -\\
ACVNet \cite{acv,xu2023accurate} & 1.2836 & 7.73 && 0.2422 & 0.737\\
Fast-ACVNet \cite{acv,xu2023accurate} & 1.1706 & 7.06 && 0.2257 & 0.740\\

% \midrule
% Urban-3D-s(no semantic) & 1.1652 & 7.02 & 13.15 & 35.39 && 1.1168 & 6.52 & 12.33 & 33.87 & \textbf{49}\\
% Urban-3D-s(ours) & 1.1190 & 6.48 & 12.46 & 34.48 && 1.0692 & 5.93 & 11.51 & 32.79 & \textbf{49}\\
\midrule
SemStereo* (ours) & 0.9956 & 5.00 && \textbf{0.2236} & \textbf{0.731}\\
SemStereo (ours) & \textbf{0.9582} & \textbf{4.58} && - & -\\
\bottomrule
\end{tabular}
\end{table*}

%% file: tables/gen.tex
\begin{table*}[!t]
\caption{\label{tab: 5}Quantitative comparison of generalization performance across cities on the US3D test set. \textbf{Bold}: Best.} % ~\cite{US3D, IADF_GRSM_201903}
\fontsize{9}{11}\selectfont
\centering
\begin{tabular}{ccccccccccccc}
\toprule
\multirow{2}{*}{Method}& \multicolumn{2}{c}{Zero-shot} && \multicolumn{2}{c}{50 Pairs Fine-tuning} && \multicolumn{2}{c}{500 Pairs Fine-tuning}\\
\cline{2-3} \cline{5-6} \cline{8-9}
 & EPE(Pixel)$\downarrow$ & D1(\%)$\downarrow$ && EPE(Pixel)$\downarrow$ & D1(\%)$\downarrow$ && EPE(Pixel)$\downarrow$ & D1(\%)$\downarrow$\\
 % &  &  &&  &\\
\midrule
StereoNet \cite{khamis2018stereonet} & 1.6719 & 11.76 && 1.6599 & 11.53 && 1.4890 & 9.58\\
PSMNet \cite{psm} &1.5163 & \textbf{9.27} && 1.3746 & 7.33 && 1.2135 & 5.53\\
GwcNet \cite{gwc} & 1.5342 & 9.32 && 1.3494 & 7.15 && 1.2467 & 5.85\\
HMSMNet \cite{he2022hmsm} & 1.5271 & 9.41 && 1.3293 & 6.95 && 1.1279 & 5.04\\
IGEV-Stereo \cite{igev} & 1.5120 & 9.91
 && 1.3659 & 8.71 && 1.2004 & 5.71\\
 ACVNet \cite{acv,xu2023accurate} & 1.5647 & 9.36 && 1.4070 & 7.55 && 1.3500 & 6.77\\
Fast-ACVNet \cite{acv,xu2023accurate} & 1.6132 & 11.13 && 1.4134 & 8.37 && 1.1753 & 5.78\\
% \midrule
% Urban-3D-s(ours) & 1.3904 & 7.98 && 1.1574 & 5.50 & \textbf{50}\\
%S3GNet-M(ours) & 1.0930 & 6.22 && 1.1507 & 5.43\\ 
SemStereo (ours) & \textbf{1.4996} & 9.70 && \textbf{1.3206} & \textbf{6.79} && \textbf{1.1002} & \textbf{4.54}\\
\bottomrule
\end{tabular}
\end{table*}

%% file: figs/5.tex
% Use figure* for multi-column figure

\begin{figure*}[!t]
\centering

% \begin{subfigure}{\linewidth}
%     \includegraphics[width=0.75\linewidth]{figs/2.png}
%     \subcaption*{\quad Left \quad \quad \quad \quad Gound truth \quad \quad \quad StereoNet \quad \quad \quad \quad PSMNet \quad \quad \quad \quad GwcNet \quad}
%     \label{fig:subfig5}
%   \end{subfigure}
%   \hfill
%   \begin{subfigure}{\linewidth}
%     \includegraphics[width=0.75\linewidth]{figs/5-2.png}
%     \subcaption*{\quad \quad ACVNet  \quad \quad \quad FastACVNet  \quad \quad \quad HMSMNet   \quad \quad \quad IGEVStereo \quad \quad \quad Urban-3D \quad \quad}
%     \label{fig:subfig6}
%   \end{subfigure}
  \includegraphics[width=0.9\linewidth]{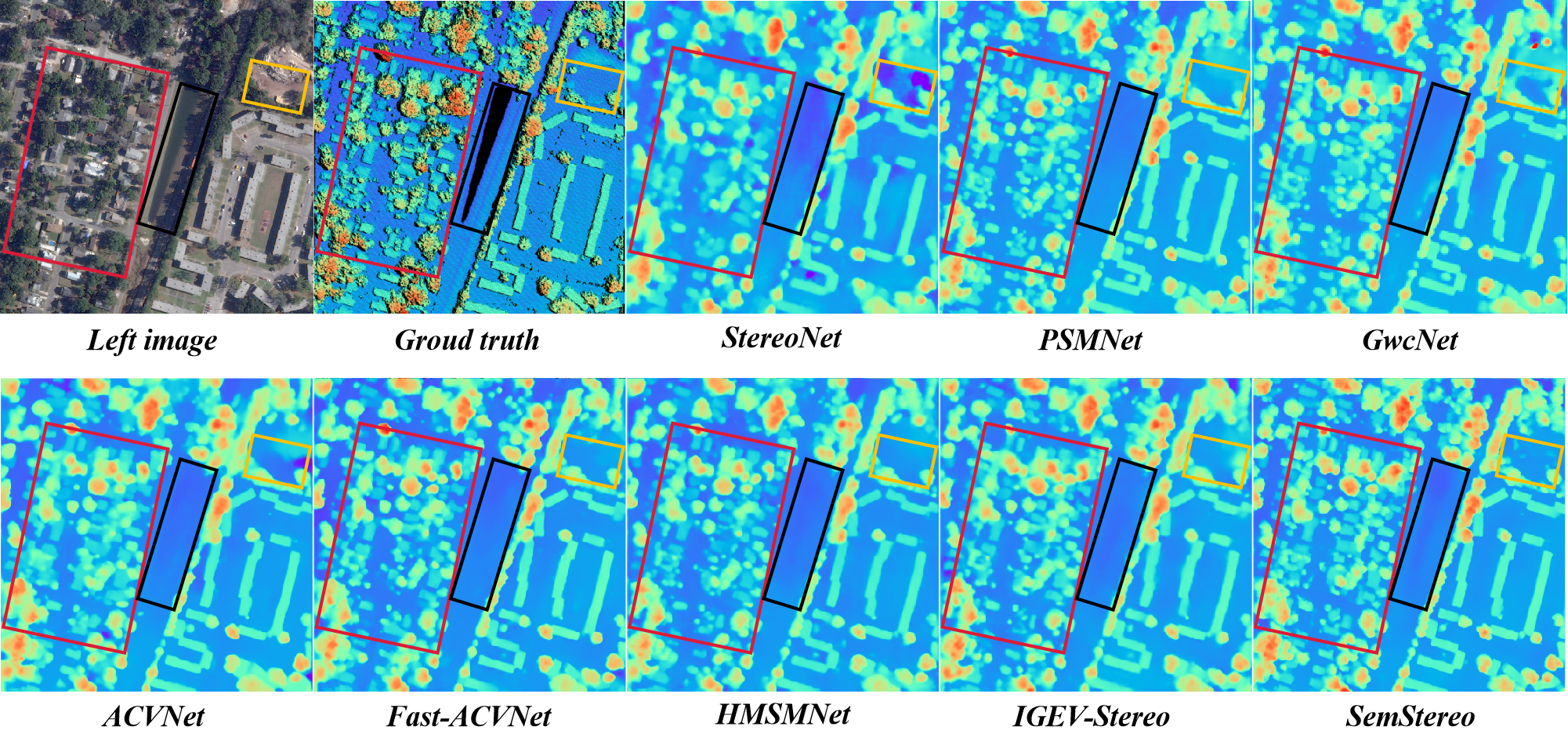}
\caption{\label{fig:5}
Qualitative comparison of results with other state-of-the-art models on the US3D test set. Red box area: Our SemStereo achieves clearer boundaries in dense buildings; Black box area: Our model has clearer boundaries even for areas without disparity labels; Yellow box area: The prediction of our model preserves more details.}
\end{figure*}
%Black, yellow, and red rectangular areas compare the performance of occluded areas, textureless areas, and borders respectively.

%% file: tables/semantic.tex
\begin{table*}[!t]
\caption{\label{tab: 6}Quantitative comparison of semantic accuracy between our SemStereo and state-of-the-art models on the US3D test set. \dag
: The results are obtained from the official declaration. * : Without stereo matching supervision. \textbf{Bold}: Best.} % *: The former represents the inference time only for the part required for the task, and the latter represents the inference time of the entire model.  ~\cite{US3D, IADF_GRSM_201903}
\fontsize{9}{10}\selectfont
\centering
\begin{tabular}{cccccccccccc}
\toprule
 \multirow{3}{*}{Method} & \multirow{3}{*}{PA(\%)$\uparrow$} & \multirow{3}{*}{mIOU(\%)$\uparrow$} & \multicolumn{5}{c}{IOU Per Class(\%)$\uparrow$}\\
 \cline{4-8}
 &  &  & \multirow{2}{*}{Ground} & \multirow{2}{*}{Trees} & Building & \multirow{2}{*}{Water} & Bridge/ \\
  &  &  &  & & Roof && Elevated Road\\
\midrule
FCN-8s \cite{long2015fully} & 88.32 & 58.89 & 86.87 & 57.62 & 75.65 & 46.59 & 27.73\\
UNet \cite{ronneberger2015u} & 90.76 & 65.98 & 86.11 & 64.04 & 81.93 & 57.53 & 40.27\\
DeepLabV3 \cite{chen2017rethinking} & 92.00 & 66.53 & 87.62 & 67.89 & 84.75 & 50.20 & 42.19\\
PSPNet \cite{zhao2017pyramid} & 91.33 & 67.06 & 86.74 & 65.12 & 82.83 & 52.80 & 47.83\\
SegFormer \cite{xie2021segformer} & 90.45 & 63.60 & 85.57 & 63.37 & 81.08 & 49.35 & 38.65\\
S$^2$Net\dag \cite{liao2023s} & - & 69.10 & 83.67 & 66.82 & 79.92 & 80.38 & 34.68\\
S$^3$Net\dag \cite{yang2024s3net} & - & 67.39 & 81.94 & 66.39 & 73.45 & 79.23 & 35.96\\
% \midrule
% Urban-3D-s(no stereo) & 92.14 & 65.08 & 87.91 & 57.77 & 85.19 & 49.98 & 44.53 & 23\\
% Urban-3D-s(ours) & 93.08 & 73.35 & 89.32 & 71.25 & 85.98 & 65.30 & 54.92 & 23\\
\midrule
SemStereo* (ours) & 92.99 & 67.57 & 89.08 & 70.60 & 87.42 & 48.37 & 42.38\\
SemStereo (ours) & \textbf{94.13} & \textbf{77.02} & \textbf{90.84} & \textbf{74.63} & \textbf{88.30} & \textbf{68.94} & \textbf{62.37}\\
\bottomrule
\end{tabular}
\end{table*}
%The upper half denotes stereo methods, while the lower denotes semantic stereo methods.%long2015fully, chen2017rethinking, zhao2017pyramid, xie2021segformer

%% file: figs/6.tex
% Use figure* for multi-column figure

\begin{figure}[!t]
\centering
\includegraphics[width=\linewidth]{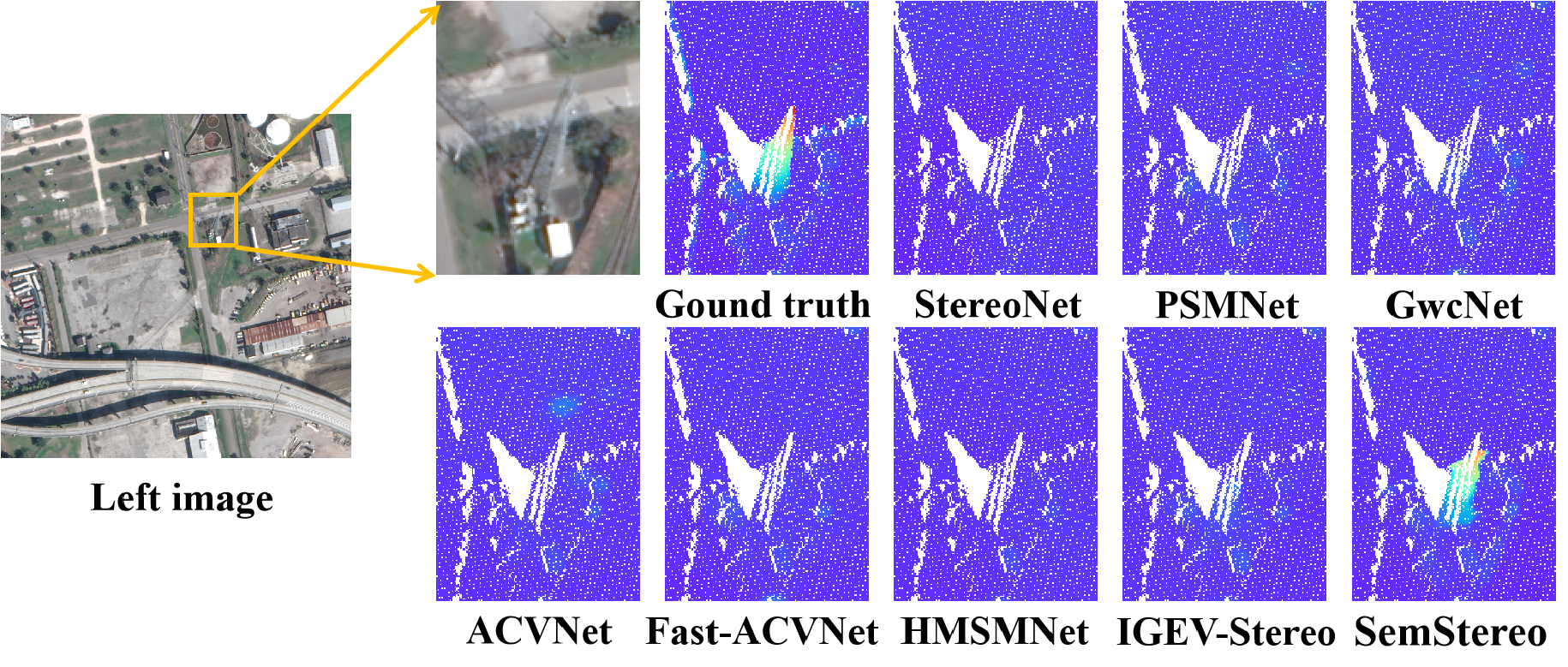}
\caption{\label{fig:6}
Qualitative comparison of results for local details with other state-of-the-art models on the US3D test set. Only our SemStereo can effectively estimate the disparities of the signal tower and its surrounding small-scale object.} %White region: Occluded region.
\end{figure}

%% file: figs/13.tex
\begin{figure*}[!t]
\centering
\includegraphics[width=\linewidth]{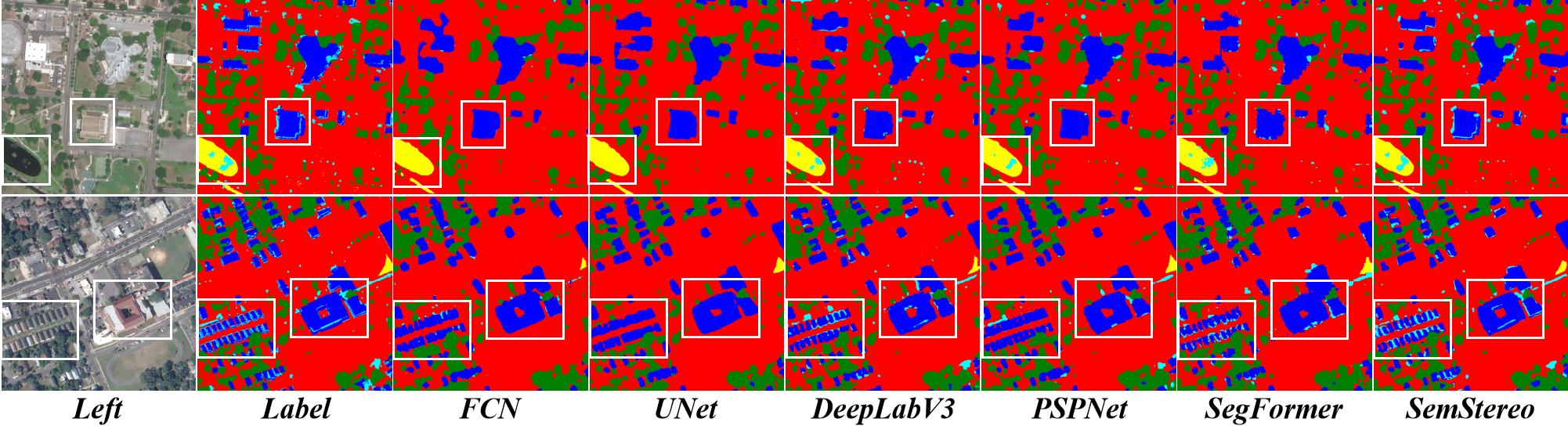}
\caption{\label{fig:12}
Qualitative comparison of semantic results with other classic state-of-the-art models on the US3D test set. Our SemStereo delivers clearer boundaries both in dense building clusters and on large objects, while also preserving more details.
} %
\end{figure*}

%% file: sec/05_conclusion.tex
\section{Conclusion}
\label{sec:conclusion}

%In this work, we observe that disparities of objects with the same category usually have close values in the realm of remote sensing showing stronger consistency than street view scenes. To fully leverage this inter-task consistency, we propose a Semantic-Guided Cascade framework for more feature consistency, a Semantic Selective Refinement branch for intra-class disparity consistency, and a semi-/self-supervised method for Left-Right Semantic Consistency.
% \textcolor{blue}{
In this work, we advocate for a deeper exploration of the complementary relationship between semantic segmentation and stereo matching in remote sensing scenes. We introduce an implicit method, SGC, which enhances feature sharing between these tasks. Additionally, we propose two explicit constraints: Semantic-Selective Refinement (SSR) to leverage disparity consistency within the same semantic category, and Left-Right Semantic Consistency (LRSC) to ensure semantic consistency across views.
Our experiments on the US3D and WHU datasets demonstrate the state-of-the-art performance of SemStereo. Ablation studies validate the effectiveness of the SGC, SSR, and LRSC modules, highlighting the mutual benefits of semantic and disparity information. We also find that semantic instances exhibit a closer relationship with disparities compared to semantic categories, and this relationship can be extended to more general scenarios. 
Future research will focus on explicitly modelling these. Furthermore, our method is also applicable to multi-view stereo with minor changes and we also plan to extend it to more applications in the future. 